\documentclass[journal]{IEEEtran}
\usepackage{amsmath,amsfonts}
\usepackage{array}
\usepackage{color}
\usepackage{bm}
\usepackage{verbatim}
\usepackage{graphicx}
\newcommand{\ctext}[1]{\raise0.2ex\hbox{\textcircled{\scriptsize{#1}}}}
\def\BibTeX{{\rm B\kern-.05em{\sc i\kern-.025em b}\kern-.08em
    T\kern-.1667em\lower.7ex\hbox{E}\kern-.125emX}}
\begin{document}
\title{Design and Validation of a Telescopic Chassis-Assisted Stair-Climbing Mechanism for Powered Wheelchairs
}
\author{Yasuhiro Fukuoka$^{1}$, Haruya Shirado$^{1}$, Yuto Nakamura$^{1}$, Haruto Okabe$^{1}$, 
Ahmad Najmuddin Ibrahim$^{2}$, \\Takahiro Fukui$^{3}$
\vspace{-5mm}
\thanks{
$^{1}$Graduate School of Science and Engineering, Ibaraki University, Japan, 
$^{2}$Faculty of Manufacturing and Mechatronic Engineering Technology, 
Universiti Malaysia Pahang Al Sultan Abdullah, Malaysia, 
$^{3}$Department of Engineering, Faculty of Engineering, Tokyo Polytechnic University, Japan}
}
\markboth{}%
{Yasuhiro Fukuoka, Haruto Okabe, Yuto Nakamura, Haruya Shirado,  Ahmad Najmuddin Ibrahim}

\maketitle

\begin{abstract}
We propose a simple stair traversal mechanism that augments a
four-wheel wheelchair with an auxiliary telescopic chassis actuator. The proposed mechanism provides the following key contributions during stair
ascent: (A) enhanced safety by ensuring that either the front or rear
wheels remain stationary at all times; (B) primary reliance of the required force and mechanical work on the telescopic chassis rather than the wheels; (C) 
insensitivity of the required force and mechanical work to stair inclination; and
(D) reduced risk of wheel slip on steeper stairs.
In addition, the telescopic chassis is expected to improve level-ground mobility
by enabling tight turning with the shortest wheelbase and enhancing stability
with the longest wheelbase.

\end{abstract}


\section{Introduction}
\label{Introduction}
Although stairs can act as barriers that restrict 
wheelchair users' access to certain destinations, situations in which 
stair climbing is unavoidable are relatively infrequent because 
elevators and ramps are often available. Therefore, climbing speed is of secondary importance, whereas safety should be prioritized because falls on stairs can result in severe injuries. The ability to traverse stairs safely, even at a slow speed, can substantially expand users' range of activities.
\par
Many powered wheelchairs capable of ascending stairs without assistance from 
caregivers have been developed over the past several decades.
However, no such wheelchair has yet achieved widespread adoption in general use.
One of the contributing factors is believed to be the excessive mechanical complexity of these systems.
%
\begin{figure}[t]
  \begin{center}
 \scalebox{0.23}{\includegraphics{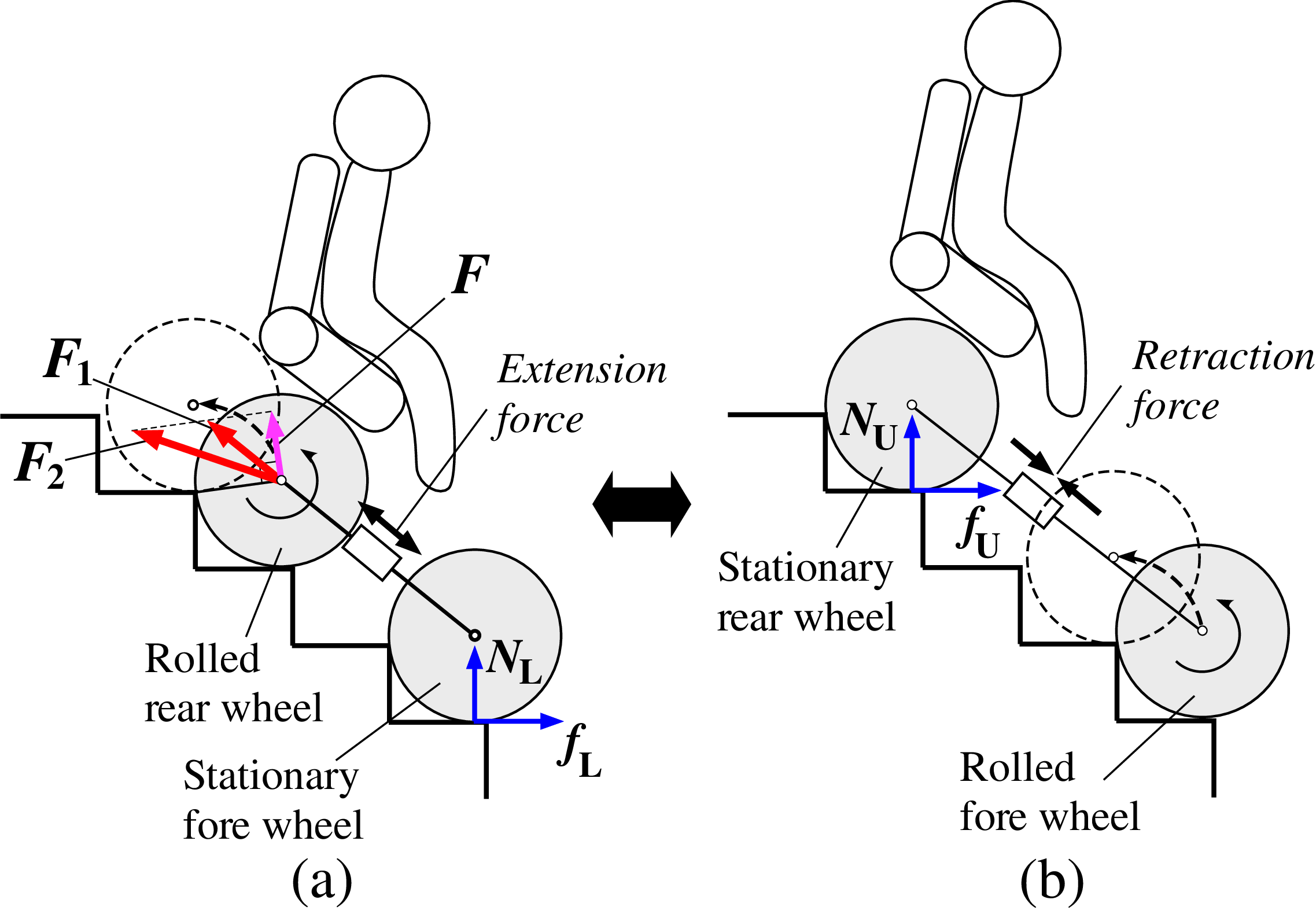}}
  \end{center}
\vspace{-3mm}
\caption{Overview of the stair ascent mechanism.}
\vspace{-6mm}
  \label{fig:diagram_climbing_stairs}
\end{figure}
Accordingly, our study aims to develop a simple mechanism that enables stair traversal while minimizing any degradation in level-ground mobility performance.
As illustrated in Fig. \ref{fig:diagram_climbing_stairs}(a) and (b), we 
propose a mobility mechanism for powered wheelchairs that enables safe 
stair ascent and descent by assisting a four-wheel configuration with a single additional chassis extension-retraction actuator. The 
proposed mechanism is characterized by its simplicity in mechanical structure, control strategy, and user operation.
As in commercially available stair-climbing wheelchairs \cite{Topchair:2026,Scewo:2026,Xsto:2026,Arthanat:2012}, the proposed system's stair ascent is performed in a backward-facing direction, whereas descent is carried out facing forward.
Each pair of front and rear wheels is driven by a single motor through self-locking worm gears, while chassis extension and retraction is achieved using a geared-motor via a ball-screw mechanism.
Specifically, stair ascent is performed as follows. In the configuration 
shown in Fig. \ref{fig:diagram_climbing_stairs}(a), the lower (front) wheels are held stationary on the stair tread by the self-locking mechanism. When the chassis is extended, the upper (rear) wheels are rolled upward about the stair edge.
After the upper wheels have landed on the next stair tread, in the 
configuration shown in Fig. \ref{fig:diagram_climbing_stairs}(b), the 
upper wheels are held stationary, and when the chassis is retracted, the lower wheels are rolled upward about the stair 
edge. By repeatedly performing the motions illustrated in Fig. 
\ref{fig:diagram_climbing_stairs}(a) and (b), the wheelchair ascends the stairs 
step by step. 
The entire motion sequence can be observed in https://youtu.be/2KtJypNziDs.
Stair descent is achieved by performing the same sequence with the roles of the front and rear wheels exchanged.
\par
The main contributions of the proposed mechanism are summarized as follows.
\begin{itemize}
\item[(A)] High safety is ensured by maintaining either the front or rear wheels in a stationary state during stair traversal.
\item[(B)] Required force and mechanical work for stair ascent rely primarily 
on the telescopic chassis rather than the wheels. 
\item[(C)] Required chassis force and total mechanical work are largely 
insensitive to stair inclination.
\item[(D)] Increased stair steepness improves traction of the stationary wheels, enhancing stability.
\item[(E)] A changeable wheelbase enabled by the telescopic chassis improves level-ground mobility.
\end{itemize}
Contribution (A) addresses an intuitive yet essential safety aspect.
Stopping one wheel pair at a time prevents simultaneous slipping of all four wheels at the stair edge.
\par
Contributions (B)-(D) constitute the novel contributions of this study.
Contribution (B) arises from the fact that the wheels can be rolled 
about the stair edge by the chassis extension-retraction force as described above.
Experimental results of climbing stairs with gentle, moderate, and steep inclination angles demonstrate that the wheel motors are
required to output only a very small torque, which is sufficient to roll the
self-locking wheels in coordination with the chassis extension and retraction
motion. 
Moreover, experiments conducted on the three stairs demonstrate that,
because the mechanical power required for the telescopic chassis is much
larger than that required for the wheels, the contribution of the
telescopic chassis to the total mechanical work exceeds 94\%.
\par
Reducing reliance on wheel-driven actuation for stair ascent offers significant advantages.
As illustrated in Figs. \ref{fig:diagram_climbing_stairs}(a) and (b), in the proposed wheelchair, the wheel radius is designed to be larger than the 
riser height of typical outdoor and public stairs, while remaining smaller than 
the tread depth so that the wheels can remain stationary on the treads.
Although wheels larger than the riser height can theoretically climb stairs using wheel torque alone, doing so with a heavy occupant requires excessively high wheel torque. 
This approach has several drawbacks. First, the required torque may 
exceed the rated torque of motors commonly used in powered wheelchairs, 
increasing energy consumption and the risk of motor overload. Second, 
steep stairs require a very high coefficient of static friction between the wheels and stair edges to avoid slipping. Third, the large tangential forces applied to the stair edges damage the stairs. Finally, the large wheel torque generates a reaction torque on the wheelchair body, increasing the risk of tip-over, particularly on steeper stairs.
However, the proposed wheelchair effectively avoids the 
concerns because of contribution (B). 
\par
In many conventional powered wheelchairs that climb stairs by lifting 
the wheels using some form of assistive mechanism 
\cite{Lawn:2003,Quaglia:2011,Yamamoto:2012,Nakajima:2020,Yuan:2005,Chocoteco:2017}, the wheel radius is significantly smaller than the stair riser height.
However, the small wheels tend to prevent progress of the vehicle over bumpy terrain.
In addition, small wheels are prone to becoming trapped in recessed gaps in the ground
 as well as in the space beneath stair risers, particularly on stairs without a vertical riser or with a
protruding stair nosing.
Thus, when priority is given to usual level-ground mobility, larger wheel radii are desirable.
\par
Regarding the second novel contribution (C), it is generally expected that 
steeper stairs require larger actuation forces and mechanical work; however, this is not the case for the proposed mechanism.
During stair ascent, the largest chassis extension-retraction force is required when the rear (upper) and front (lower) wheels lift off the stair tread, as shown in Figs. \ref{fig:diagram_climbing_stairs}(a) and (b), respectively.
Because the steeper configuration shown in Figs. 
\ref{fig:diagram_climbing_stairs} results in a larger gravitational 
component acting along the body axis, a correspondingly larger 
extension-retraction force is required to counteract gravity. However, the chassis extension-retraction
force required to rotate the wheels about the stair edge against gravity 
is, in fact, smaller than in less steep configurations.
Specifically, for the same rolling force $F$ in 
Fig.\ref{fig:diagram_climbing_stairs}(a) acting on the upper wheels about 
the stair edge, the required chassis extension force $F_1$ is smaller than $F_2$ required in gentle inclination.
Our experiments demonstrate that the overall maximum chassis actuation force
required does not vary significantly with changes in the inclination of the
three tested stairs. Moreover, the maximum force was only approximately 10 \% of 
the rated actuation force. 
The experimental results further indicate that the total
mechanical work normalized by vertical displacement decreases as the stair
steepness increases.
\par
For the third novel contribution (D), chassis extension and retraction 
generate backward forces on the stationary lower and upper wheels, 
denoted as $f_{\rm{L}}$ and $f_{\rm{U}}$, respectively (Fig. 
\ref{fig:diagram_climbing_stairs}(a), (b)). With increasing body 
inclination, the actuation force shifts toward the vertical direction, 
reducing the horizontal forces $f_{\rm{L}}$ and $f_{\rm{U}}$ while 
increasing the normal forces $N_{\rm{L}}$ and $N_{\rm{U}}$. As a result, 
analysis and experiments show that wheel slip is less likely on steeper stairs.
\par
Contribution (E) represents an additional benefit for usual level-ground operation.
Although turning is outside the scope of this paper, since chassis retraction 
enables a substantially reduced wheelbase, 
a smaller minimum turning radius can be achieved than that of conventional powered wheelchairs.
In addition, extending the chassis on bumpy terrain or slopes increases the wheelbase, reducing the probability of forward and backward tipping. 
Thus, the telescopic chassis may further enhance the performance of 
conventional wheelchairs. 
\section{Existing stair-climbing powered wheelchairs}
\label{Related works}
Many stair-climbing mechanisms have been extensively reviewed in the literature 
\cite{Seo:2023}. 
Even focusing solely on powered wheelchairs, a wide range of stair-climbing mechanisms has been reviewed in the literature \cite{Sundaram:2017,Verma:2022}.
Although enabling wheel-independent stair climbing through the active telescopic chassis is one of the contributions of the proposed mechanism, as described in Contribution (B), existing stair-climbing wheelchairs achieve the same objective using a variety of different mechanisms.
\par
The most widely developed and commercially available approach is a mechanism 
that switches between wheels for level-ground locomotion and crawlers for stair 
traversal \cite{Topchair:2026,Scewo:2026,Xsto:2026,Tao:2016,Sharma:2022}.
However, each crawler belt has only two or three contact points with the stair edges during stair traversal, whereas our mechanism ensures static stability by keeping either the front or rear wheels on a tread at all times, as described in Contribution (A).
On steeper stairs, the crawler requires higher torque and power and 
becomes more susceptible to slipping. These characteristics are contrary to Contributions (C) and (D). 
In addition, because wheel-crawler wheelchairs that use four wheels for level-ground locomotion \cite{Topchair:2026,Xsto:2026,Tao:2016} cannot significantly change their wheelbase, they cannot provide the benefits as described in Contribution (E), such as extremely tight turning with a shortened body and enhanced fore-aft stability on uneven terrain with an extended body.
Furthermore, smooth switching between the wheel and crawler modes requires 
complicated control at stair entry and exit. 
Moreover, because the 
crawler length is limited, gentle stairs with a large tread depth may 
reduce the number of contact points between the crawler and the stair edges.
Even if there are two contact points, a forward or rearward concentration of those contact points can increase the risk of tipping.
Safety can only be guaranteed on stairs where three contact points can be maintained for a sufficiently long period.
Furthermore, in the event of slipping, the flat bottom surface of crawlers may 
cause the wheelchair to slide all the way down the stairs, and asymmetric slipping of a single crawler can induce yaw rotation and lead to lateral tipping.
\par
Most stair-climbing mechanisms that do not employ crawlers rely on some form of wheel-lifting strategy.
A representative example is a wheel-cluster mechanism in which multiple wheels are arranged in a planetary configuration on a common axis: conventional locomotion is achieved through wheel rotation, whereas stair climbing is accomplished via revolution about the central axis.
Wheelchairs employing a two-wheel cluster mechanism on each side 
require either human assistance, the 
use of a handrail \cite{Arthanat:2012}, or active inverted-pendulum control to maintain 
stability during stair ascent \cite{Onozuka:2020}.
Employing front and rear pairs of left-right wheel-cluster 
mechanisms improves overall stability 
\cite{Lawn:2003,Quaglia:2011,Yamamoto:2012}. 
However, these mechanisms does not provide advantages 
similar to Contributions (A), (C) and (D).
In addition, the use of multi-component clusters increases mechanical and control 
complexity, and the small wheels forming the clusters may impede 
efficient level-ground locomotion, as noted in Sec. \ref{Introduction}.
Although the wheelchair \cite{Yamamoto:2012} also employs a passive telescopic chassis, it cannot assist wheel progression.
\par
Some stair-climbing mechanisms employ multiple links to lift small wheels, 
enabling stair ascent in a manner similar to that of legged robots 
\cite{Nakajima:2020,Yuan:2005,Chocoteco:2017}. 
Although these stair-climbing mechanisms can achieve very high static 
stability, potentially surpassing Contribution (A), they require 
multiple actuators and movable links to lift the wheels and do not 
provide the advantages similar to Contributions (C)-(E).
In addition, the small wheel size may degrade level-ground mobility performance.
\section{Static analysis}
\label{Static analysis}
\subsection{Equilibrium analysis at the critical moments of stair ascent}
\label{Equilibrium analysis at the critical moment of stair ascent}
\begin{figure}[t]
\vspace{-2mm}
  \begin{center}
\scalebox{0.3}{\includegraphics{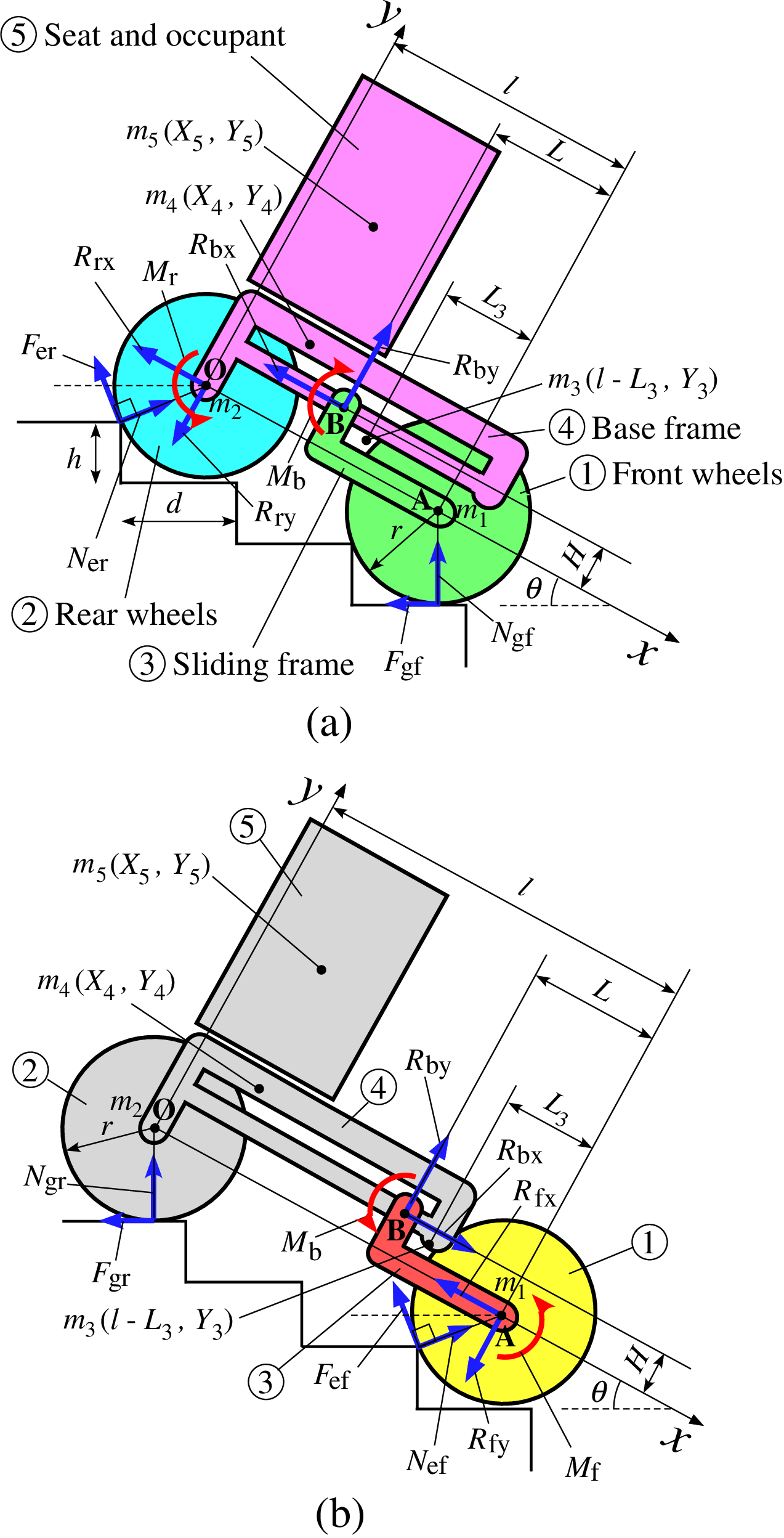}}
  \end{center}
\vspace{-4mm}
\caption{Free-body diagrams of the wheelchair during stair ascent. }
\vspace{-2mm}
  \label{fig:free_body_diagram}
\end{figure}

Since the objective of this study is to achieve gradual stair ascent to ensure safety, static equilibrium in a two-dimensional plane is considered.
Figs \ref{fig:free_body_diagram}(a) and (b) are the free-body diagrams 
corresponding to Figs. \ref{fig:diagram_climbing_stairs}(a) and (b), respectively.
Figs. \ref{fig:free_body_diagram}(a) and (b) depict the moments at which the
upper and lower wheels lift off from the stair tread to climb one step, respectively.
Since these moments require the largest chassis telescopic force during 
the chassis extension and retraction motions, Figs. \ref{fig:free_body_diagram}(a) and (b) are selected as the focus of the analysis.
$x$- and $y$-axes are defined as shown in the figures. 
The wheelchair is composed of five units: the \ctext{1} front wheels, \ctext{2} 
rear 
wheels, \ctext{3} sliding frame, \ctext{4} base frame, and \ctext{5} the seat and occupant. \ctext{4} and \ctext{5} are treated as a 
single rigid body, except when the seat is slid forward or backward. 
Let $m_1$ through $m_5$ denote the masses of the five respective units. 
The center-of-mass coordinates of $m_3$, $m_4$, and $m_5$ are 
given by ($l-L_3$, $Y_3$), ($X_4$, $Y_4$), and ($X_5$, $Y_5$), respectively.
The variable $l$ is the wheelbase and changes with the chassis extension length. 
The radii of both the front and rear wheels are denoted by $r$. $\theta$ represents the body inclination angle.
%
%
\par
In Fig. \ref{fig:free_body_diagram}(a), \ctext{1} and \ctext{3} are regarded as a single rigid body. 
Accordingly, the vector of unknowns consisting of the forces and moments acting 
on the three differently colored units is given as
%
$
\small
\mathbf{q_e}^{\mathsf T} =
\begin{bmatrix}
R_{\rm{rx}} & R_{\rm{ry}} & R_{\rm{bx}} & R_{\rm{by}} & N_{\rm{er}} & F_{\rm{er}} & N_{\rm{gf}} & F_{\rm{gf}} & M_{\rm{r}} & M_{\rm{b}}
\end{bmatrix},
$
%
where $R_{\rm{rx}}$, $R_{\rm{ry}}$, and $M_{\rm{r}}$ are the reaction forces and 
moment applied by \ctext{4} to \ctext{2} at O; $R_{\rm{bx}}$, $R_{\rm{by}}$, and $M_{\rm{b}}$ are those applied by \ctext{3} to \ctext{4} at B.
$N_{\rm{er}}$ and $F_{\rm{er}}$ are the normal and friction forces from the stair edge acting on \ctext{2}, while $N_{\rm{gf}}$ and $F_{\rm{gf}}$ are the corresponding forces from the ground acting on \ctext{1}.
Here, $R_{\rm{bx}}$ represents the chassis extension force, and $M_{\rm{r}}$ represents the driving torque of the upper wheels.
\par
Similarly, in Fig. \ref{fig:free_body_diagram}(b), \ctext{2}, \ctext{4}, and \ctext{5} are regarded as a single rigid body. 
Accordingly, the vector of unknowns consisting of the forces and moments acting 
on the three differently colored units is given as
%
$
\small
\mathbf{q_r}^{\mathsf T} =
\begin{bmatrix}
R_{\rm{fx}} & R_{\rm{fy}} & R_{\rm{bx}} & R_{\rm{by}} & N_{\rm{ef}} & F_{\rm{ef}} & N_{\rm{gr}} & F_{\rm{gr}} & M_{\rm{f}} & M_{\rm{b}}
\end{bmatrix},
$
%
where $R_{\rm{fx}}$, $R_{\rm{fy}}$, and $M_{\rm{f}}$ are the reaction forces and 
moment applied by \ctext{3} to \ctext{1} at A; $R_{\rm{bx}}$, $R_{\rm{by}}$, and $M_{\rm{b}}$ are those applied by \ctext{3} to \ctext{4} at B.
$N_{\rm{ef}}$ and $F_{\rm{ef}}$ are the normal and friction forces from the stair edge acting on \ctext{1}, while $N_{\rm{gr}}$ and $F_{\rm{gr}}$ are the corresponding forces from the ground acting on \ctext{2}.
Here, $R_{\rm{bx}}$ represents the chassis retraction force, and $M_{\rm{f}}$ represents the driving torque of the lower wheels.
%
\begin{figure*}[t]
  \begin{center}
\scalebox{0.23}{\includegraphics{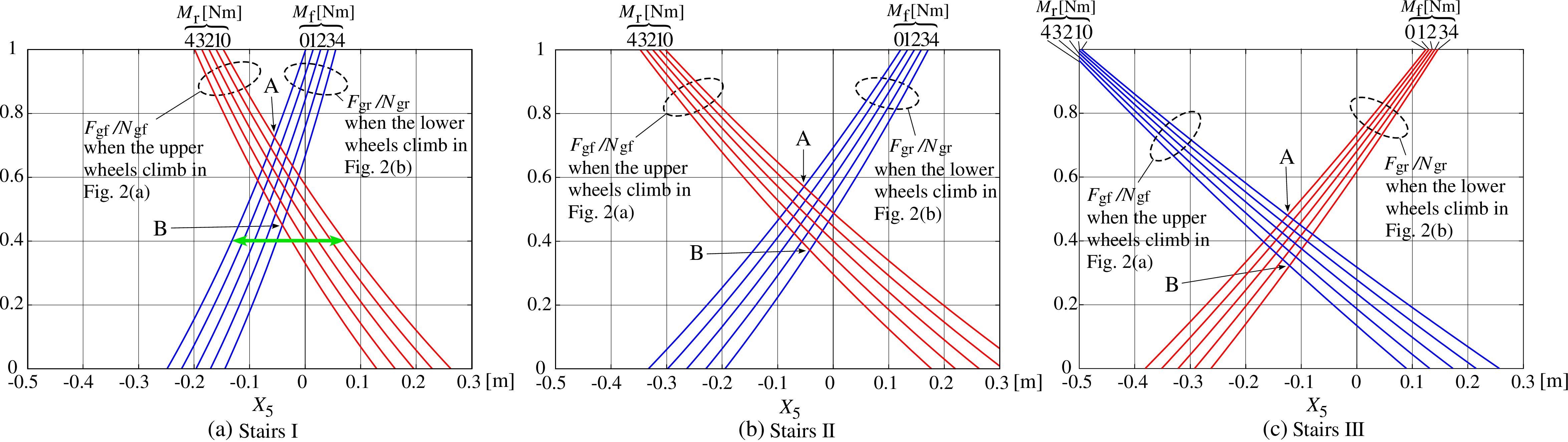}}
  \end{center}
\vspace{-4mm}
\caption{Relationship between the position of the seat and occupant unit 
$X_5$ and slip tendency for the configurations shown in Figs. \ref{fig:free_body_diagram} (a) and (b).}
\vspace{-2mm}
  \label{fig:simulation_result}
\end{figure*}
\subsection{Parameter values}
\label{Parameter values}
The analytical model in Figs. \ref{fig:free_body_diagram}(a) 
and (b) and the prototype wheelchair (Sec. \ref{Prototype wheelchair}) are 
assumed to be at one-half scale in dimensions and one-tenth in mass relative to 
a future full-scale powered wheelchair. The parameters used in the analysis are as follows:
the wheel radius is $r = 107$ mm, the mass of the five units are $m_1 = m_2 = 1.0$ kg, $m_3 = 1.8$ kg, $m_4 = 6.2$ kg, $m_5 = 10.0$ kg; the geometric parameters are $L_3 = 
110$ mm, $Y_3 = 25$ mm, $X_4 = 60$ mm, $Y_4 = 57$ mm, $Y_5 = 270$ mm, $L = 210$ 
mm, $H = 50$ mm. 
The masses and center-of-mass locations were determined based on the prototype wheelchair.
The appropriate value of the seat position $X_5$ varies depending on the stair 
inclination (see Sec. \ref{Analysis and discussion} for details).
Table \ref{tab1} lists the three types of stairs used in the analysis in Sec. 
\ref{Analysis and discussion} and the experiments in Sec. 
\ref{Experimantal results} together with the corresponding values of the 
riser height $h$, tread depth $d$.
Stair I, Stair II, and Stair III represent gentle, standard, and steep stairs, respectively, as commonly found in outdoor environments and public facilities.
%
\begin{table}[b]
\vspace{-4mm}
\begin{center}
  \caption{Stair types and parameters used in analysis and experiments}
 \label{tab1}
\vspace{-2mm}
\begin{tabular}{llll} \hline
      & Stair I   & Stair II  & Stair III  \\ \hline
Stair inclination [deg]   & 24 & 32  & 40   \\ \hline
Riser height $h$ [mm]     & 75   & 85    & 95  \\  \hline
Tread depth $d$ [mm] &  170    &  135    & 115  \\ \hline
Wheelbase $l$ [mm]   &  2$r$   &  2$\sqrt{h^2 + d^2}$   & 2$\sqrt{h^2 + d^2}$  \\ 
at Fig. \ref{fig:free_body_diagram}(a)   & = 220  & = 319 & = 298 \\ \hline
Wheelbase $l$ [mm]   &  2$\sqrt{h^2 + d^2}$   &  3$\sqrt{h^2 + d^2}$   & 3$\sqrt{h^2 + d^2}$  \\ 
at Fig. \ref{fig:free_body_diagram}(b)   & = 372  & = 478 & = 447 \\ \hline
\end{tabular}
\end{center}
\vspace{-7mm}
\end{table}
According to the Blondel rule \cite{Neufert:2012}, a comfortable stair generally satisfies $2h + d \approx 63\,\mathrm{cm}$, and the rule is widely used as a guideline for stair design. 
The values of $h$ and $d$ in Table \ref{tab1} are determined according to the Blondel rule for each stair inclination.
The wheelbase $l$ in the configurations shown in Figs. \ref{fig:free_body_diagram}(a) and (b) is geometrically determined by the stair dimensions, as listed in Table \ref{tab1}.
In the cases of Stair II and III, the upper and lower wheels move with a 
separation of at least one step, as shown in Figs. \ref{fig:free_body_diagram}(a) and (b), to prevent wheel contact and reduce the risk of overturning on steep slopes.
As a result, the wheelbase $l$ is much longer than that for Stair I, where both the front and rear wheels can be placed on consecutive treads.
%
\begin{figure*}[t]
  \centering
  \begin{minipage}{0.55\linewidth}
    \centering
    \includegraphics[scale=0.38]{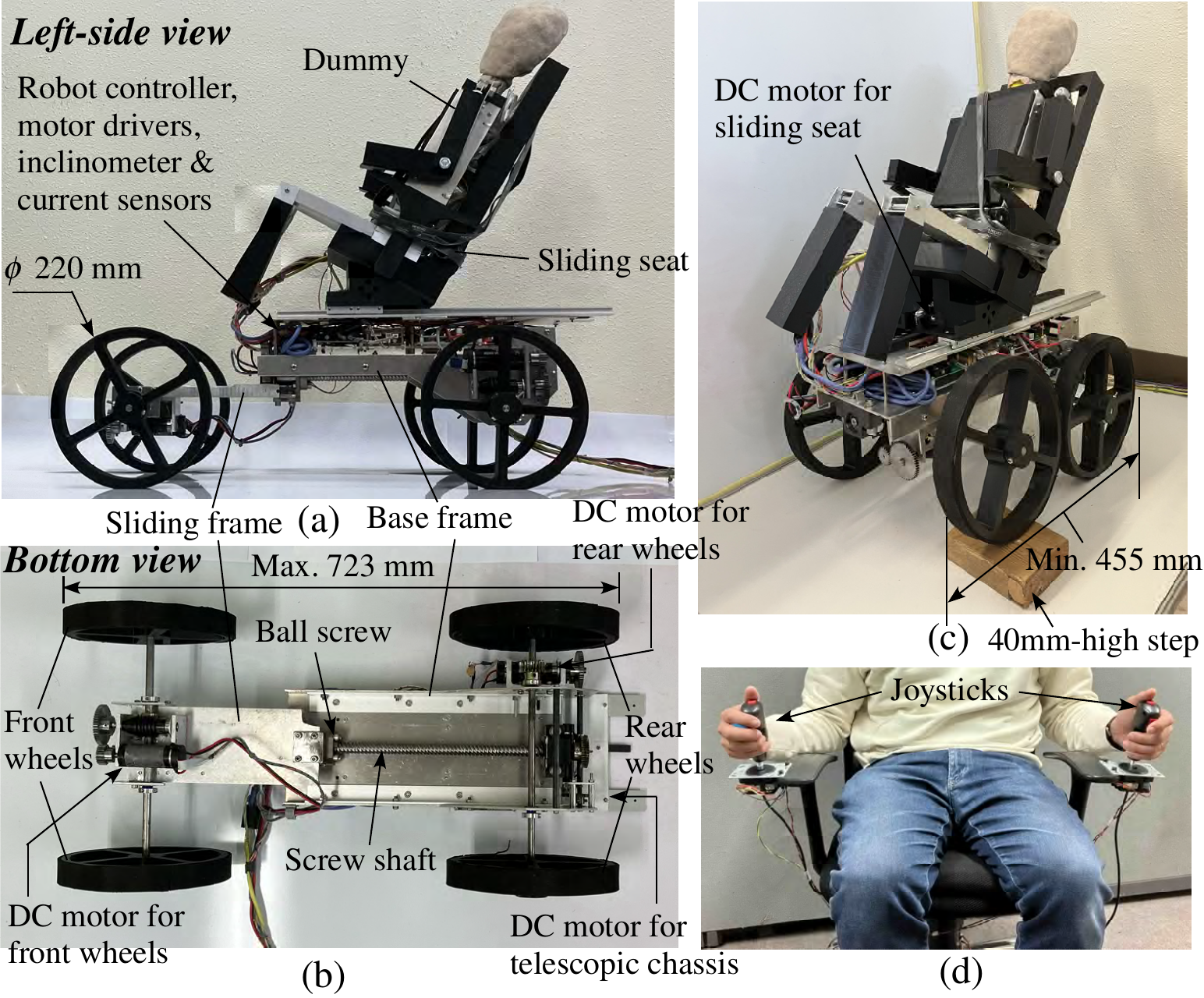}
  \caption{Prototype wheelchair and teleoperation interface.}
  \label{fig:photos}
  \end{minipage}
  \hfill
  \begin{minipage}{0.44\linewidth}
    \centering
    \includegraphics[scale=0.33]{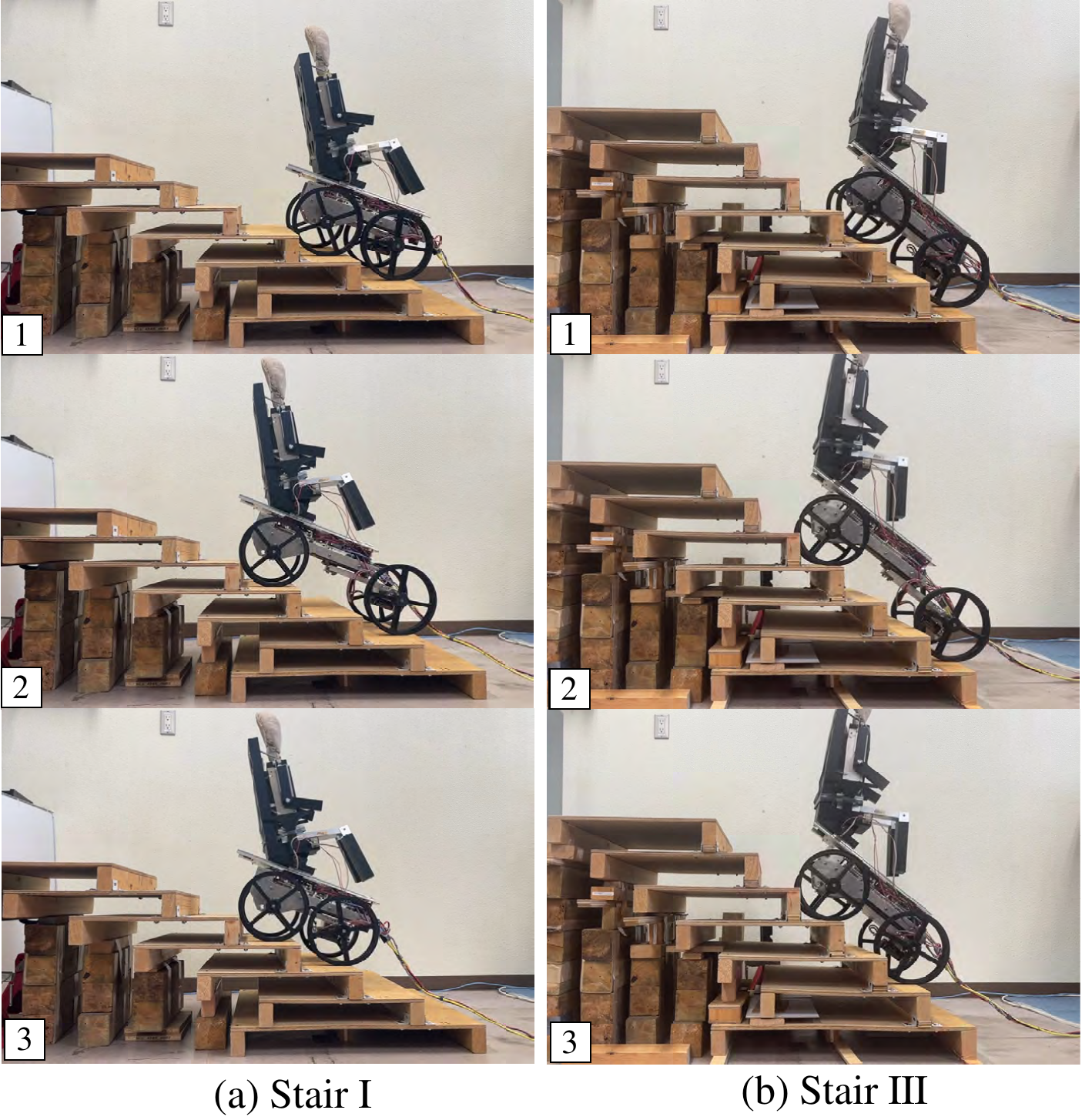}
  \caption{Snapshots when climbing stairs.}
  \label{fig:snapshots}
  \end{minipage}
  \vspace{-5mm}
\end{figure*}
\subsection{Analysis and discussion}
\label{Analysis and discussion}
Nine equilibrium equations --- consisting of force balance in two directions and 
moment balance about an arbitrary point for the three units---are obtained for 
each of the configurations shown in Figs. \ref{fig:free_body_diagram}(a) and (b).
By prescribing the applied torque $M_r$ on the rear (upper) wheels in
Fig. \ref{fig:free_body_diagram}(a) and the applied torque $M_f$ on the front
(lower) wheels in Fig. \ref{fig:free_body_diagram}(b) to values of 0, 1, 2, 3,
and 4 Nm, the number of unknowns in the vectors $\mathbf{q_e}$ and
$\mathbf{q_r}$ defined in Sec. \ref{Equilibrium analysis at the critical moment of stair ascent} can be reduced.
Under these conditions, the nine equilibrium equations can be solved 
numerically using MATLAB to obtain $\mathbf{q_e}$ and $\mathbf{q_r}$.
\par
Figs. \ref{fig:simulation_result}(a), (b), and (c) show, for Stair I, II, and 
III, respectively, the ratios $F_{\rm{gf}}/N_{\rm{gf}}$ in Fig. 
\ref{fig:free_body_diagram}(a) and $F_{\rm{gr}}/N_{\rm{gr}}$ in Fig. 
\ref{fig:free_body_diagram}(b) plotted in red and blue, respectively, as functions of the parameter $X_5$, which denotes the $x$-position of the seat and occupant.
If the ratios $F_{\rm{gf}}/N_{\rm{gf}}$ and $F_{\rm{gr}}/N_{\rm{gr}}$ are smaller than the coefficient of static friction $\mu$ between the wheels 
and the stair tread, the system can maintain static equilibrium without 
slipping. 
The ratio \(F_{\mathrm{gf}}/N_{\mathrm{gf}}\) (red plots in Fig. \ref{fig:simulation_result})
decreases with increasing \(X_5\),
because the normal force \(N_{\mathrm{gf}}\) on the stationary wheel increases
as the center of mass approaches it.
For the same reason, \(F_{\mathrm{gr}}/N_{\mathrm{gr}}\) (blue plots) decreases as 
\(X_5\) decreases.
\par
As shown in Fig. \ref{fig:simulation_result}(a), for the case $M_r = M_f = 0$, 
which represents stair ascent achieved solely through chassis extension and 
retraction without applying wheel torque, both 
$F_{\mathrm{gf}}/N_{\mathrm{gf}}$ and $F_{\mathrm{gr}}/N_{\mathrm{gr}}$ are 
equal to 0.72 when the center of mass of the occupant-carrying seat is at $X_5 = -0.057$ m (intersection 
point A in Fig. \ref{fig:simulation_result}(a)).
That is, if the coefficient of static friction satisfies $\mu > 0.72$, the 
wheelchair can climb Stair I using only chassis extension and retraction, 
while keeping the seat fixed at the position $X_{5} = -0.057$ m. 
\par
It is noteworthy that, as described in contribution (D) of Sec. 
\ref{Introduction}, increasing the stair inclination leads to a reduction in 
the vertical-axis value at the intersection point where \(F_{\mathrm{gf}}/N_{\mathrm{gf}} = F_{\mathrm{gr}}/N_{\mathrm{gr}}\).
Specifically, in Fig. \ref{fig:simulation_result}, for the case $M_r = M_f = 0$,
the vertical values of the intersection-point A are
0.72 ($X_5 = -0.057$ m) for (a) Stair I, 0.57 ($X_5 = -0.055$ m) for (b) Stair II,
and 0.48 ($X_5 = -0.126$ m) for (c) Stair III.
These results can be attributed to the fact that, as the body becomes more 
inclined, the extension-retraction force is increasingly directed vertically 
rather than horizontally. This reduces the friction forces \(F_{\mathrm{gf}}\) and \(F_{\mathrm{gr}}\) while increasing the normal forces \(N_{\mathrm{gf}}\) and \(N_{\mathrm{gr}}\) on the stationary wheels.
\par
If torques of $M_r = M_f = 4$ Nm are applied to 
the upper and lower wheels at the configurations in 
Figs. \ref{fig:free_body_diagram}(a) and (b), respectively, the intersection 
points are largely shifted downward to Point B for all stairs. This indicates 
that stairs with more slippery treads can be climbed. 
However, excessively large wheel torques $M_r$ and $M_f$ are undesirable 
for the reasons described in Sec. \ref{Introduction}.
Alternatively, without increasing $M_r$ and $M_f$, the friction ratios 
$F_{\mathrm{gf}}/N_{\mathrm{gf}}$ and $F_{\mathrm{gr}}/N_{\mathrm{gr}}$ can be 
reduced by reciprocating the seat position to modify $X_5$. 
For example, sliding the seat to positions $X_5 = 0.07$ m and -0.13 m in Figs. 
\ref{fig:free_body_diagram}(a) and (b), respectively, results in $F_{\mathrm{gf}}/N_{\mathrm{gf}} = F_{\mathrm{gr}}/N_{\mathrm{gr}} = 0.4$, enabling stair ascent even on highly slippery surfaces.
This is confirmed by the green arrows in Fig. \ref{fig:simulation_result}(a) 
when $M_r = M_f = 0$. 
\section{Prototype powered wheelchair and its operation and drive method}
\label{Prototype powered wheelchair and its operation and drive method}
\subsection{Prototype powered wheelchair}
\label{Prototype wheelchair}
Figs. \ref{fig:photos}(a)-(c) show photographs of the prototype wheelchair: (a) 
left-side view and (b) bottom view at the maximum extension and (c) view at the minimum extension. The weight of the prototype is 11.9 kg. 
In the experiments, an 8 kg anthropomorphic dummy composed of 11 movable parts, whose masses approximate those of the corresponding human body segments, was mounted on the prototype and secured with a seat belt. The parts were elastically connected to reproduce the compliance of the human body.
Owing to the chassis telescopic mechanism based on a 
ball-screw drive shown in Fig. \ref{fig:photos}(b), the wheelbase $l$ can be varied over a range of 225-503 mm.
Consequently, the overall length of the prototype is 455 mm at the minimum 
extension and 723 mm at the maximum extension.
The masses and center-of-mass locations of the five main components of the 
prototype are identical to those applied in the analytical model described in 
Sec. \ref{Static analysis}. 
The wheels consist of plastic wheels with a radius of 105 mm, wrapped with a 
5-mm-thick layer of urethane rubber. Owing to elastic compression of the 
urethane layer under load, the wheel radius becomes 107 mm on stair treads, 
consistent with that of the analytical model used in Sec. \ref{Static analysis}.
The wheel width is 30 mm. 
The static friction coefficient between the wheels and the tread surface of the handmade wooden stairs was experimentally measured as $\mu = 0.75$.
By comparison, the static friction coefficient between automobile tires and 
dry asphalt is commonly reported to be in the range of 0.8-1.2 \cite{Lorenz:2015}.
\par
As shown in Fig. \ref{fig:photos}(b), the front and rear wheels are each driven 
by a 60 W DC motor (Maxon RE30) equipped with a worm-gear reducer with a 
reduction ratio of 40:1. As a result, the rated torque and maximum angular 
velocity of both the front and rear wheels are 3.42 Nm and 23.1 rad/s, respectively.
The maximum translational speed is therefore 23.1 rad/s $\times$ 0.107 m = 2.47 
m/s.
\par
As shown in Fig. \ref{fig:photos}(b), the telescopic mechanism employs a ball screw connected to the sliding frame, which translates along a screw shaft fixed under the base frame.
The screw shaft is rotated by a DC motor identical to the wheel-drive motor via 
a 40:1 worm-gear reducer.
The ball-screw mechanism provides a maximum chassis linear speed of 73.4 mm/s and a 
rated thrust of 968 N.
In addition, the sliding frame with the front wheels can rotate 
passively about the screw shaft, enabling all four wheels to maintain 
traction even over laterally bumpy terrain, as shown in Fig. \ref{fig:photos}(c).
\par
As shown in Fig. \ref{fig:photos}(a), the seat supporting the dummy is 
mounted on a rack-and-pinion mechanism driven by a 20 W DC motor (Maxon 
RE25) with a 100:1 planetary gear reducer, allowing the seat to slide 
longitudinally to adjust its fore-aft center-of-gravity position $X_5$ in Figs. \ref{fig:free_body_diagram} and \ref{fig:simulation_result}. 
\par
Each DC motor is equipped with an 
encoder to measure angular position and velocity, from which the linear 
chassis extension length and the seat position are also derived.
The current flowing through each motor cable is measured using current 
sensors to compute the output torque.
An inclination sensor is used to measure the vehicle pitch angle.
The wheelchair is controlled by a real-time robot controller. 
Although the prototype is powered externally, an onboard battery is readily accommodated in the future full-scale design owing to the sufficiently large mass allowance of the base frame.
%
\begin{table}[b]
\vspace{-4mm}
\scriptsize
\begin{center}
  \caption{Operating modes for the prototype wheelchair}
\vspace{-3mm}
 \label{tab2}
\begin{tabular}{llll} \hline
Mode & Operation       & Operation         & Achieved motion  \\ 
     & of the left  & of the right &   \\
     & joystick & joystick  &   \\ \hline
0    & Neutral & Neutral & Set all motors to zero output   \\ \hline
1    & Forward & Forward & Move forward while keeping  \\ 
     &         &         & the current chassis length  \\  \hline
2    & Backward& Backward& Move backward while keeping  \\ 
     &         &         & the current chassis length  \\  \hline
3    & Neutral & Backward& Rear wheels backward with chassis  \\ 
     &         &         & extension (front wheels stopped) \\ \hline
4    & Backward& Neutral & Front wheels backward with chassis  \\ 
     &         &         & retraction (rear wheels stopped) \\ \hline
5    & Neutral & Forward & Front wheels forward with chassis  \\ 
     &         &         & extension (rear wheels stopped) \\ \hline
6    & Forward & Neutral & Rear wheels forward with chassis  \\ 
     &         &         & retraction (front wheels stopped) \\ \hline
7    & Leftward & Neutral & Sheet slides backward  \\ \hline
8    & Rightward & Neutral & Sheet slides forward \\ \hline
\end{tabular}
\end{center}
\vspace{-7mm}
\end{table}
\subsection{Operation and drive method}
\label{Operation and drive method}
As illustrated in Fig. \ref{fig:photos}(d), the wheelchair is teleoperated using two four-direction ON-OFF joysticks attached to the
left and right armrests of an external chair, while the operator visually monitors the prototype.
Nine operating modes are available (Table \ref{tab2}).
On level ground, locomotion is performed using Modes 1 and 2, and the chassis 
length can be adjusted using Modes 3-6.
For stair climbing and descending, Modes 3 and 4 and Modes 5 and 6 are used, respectively.
The seat is slid by Modes 7 and 8. 
\par
In Modes 1 and 2, forward and backward locomotion is achieved by controlling both the front and rear wheels at arbitrary velocities.
During stair ascent, because the four wheels are self-locked by the 
worm gears, they cannot be passively rotated even when $M_r = M_f = 0$ 
in Modes 3 and 4, corresponding to Figs. \ref{fig:free_body_diagram}(a) 
and (b). To approximately mimic passive rotation, a small constant 
torque of $M_r = M_f = 0.3$ Nm is applied as the command value 
throughout all the stair-ascent experiments.
In Mode 3, the chassis extension velocity (positive in the extension direction) is regulated by P control based on the reference velocity given as
\begin{equation}\label{desired_velocity}
v_d = V_d t^2\ \ (0 \le t \le 1 s), \ v_d = V_d\ \ (t > 1 s).
\end{equation}
Using this equation, the extension velocity is initially increased gradually to minimize acceleration effects, and then the chassis is extended 
at a constant velocity $V_d$. The value of $V_d$ as well as the P gain are selected empirically.
In Mode 4, a reference velocity obtained by negating Eq. (\ref{desired_velocity}) is used for the chassis contraction.
During stair ascent, Mode 3 and 4 are executed alternately.
During stair descent, Mode 5 and 6 are executed alternately, where the same 
driving principles are employed as Mode 3 and 4, respectively, with the roles of the front and rear wheels exchanged.
\section{Experimental results}
\label{Experimantal results}
\begin{figure*}[t]
  \centering
\scalebox{0.345}{\includegraphics{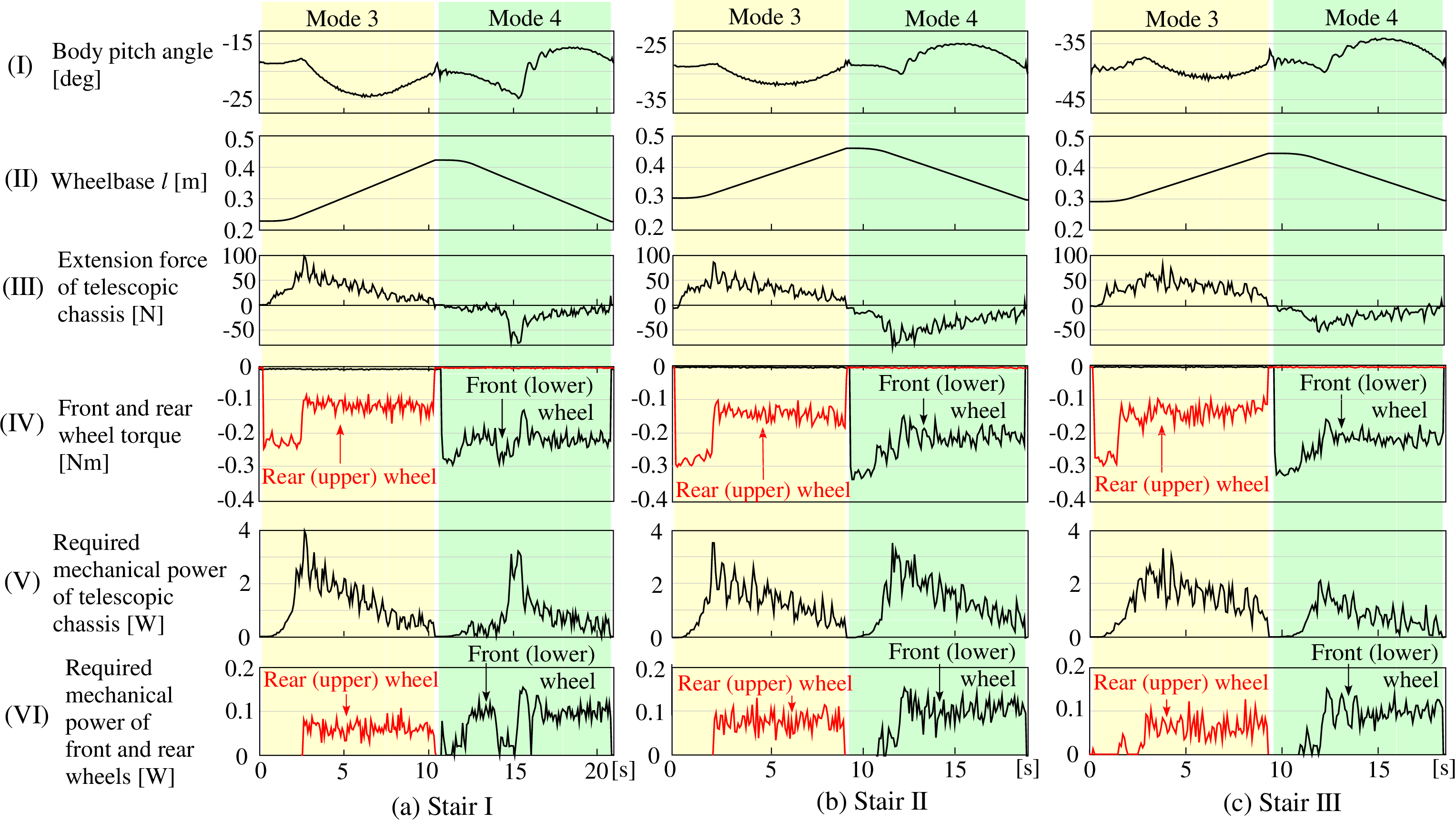}}
\vspace{-2mm}
  \caption{Experimental results of stair-climbing.}
\label{fig:graph}
\vspace{-5mm}
\end{figure*}
\begin{figure}[t]
  \centering
\scalebox{0.53}{\includegraphics{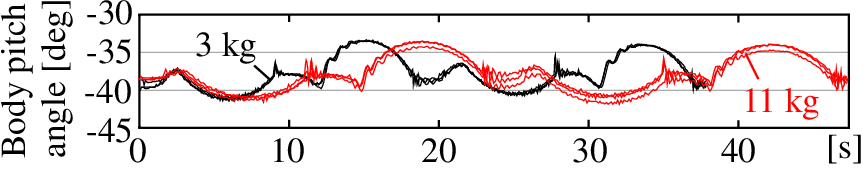}}
\vspace{-2mm}
\hspace{10mm}
  \caption{Results of climbing Stair III with light and heavy dummies.}
\label{fig:dummy-3kg-and-11kg}
\end{figure}
\begin{figure}[t]
  \centering
\scalebox{0.51}{\includegraphics{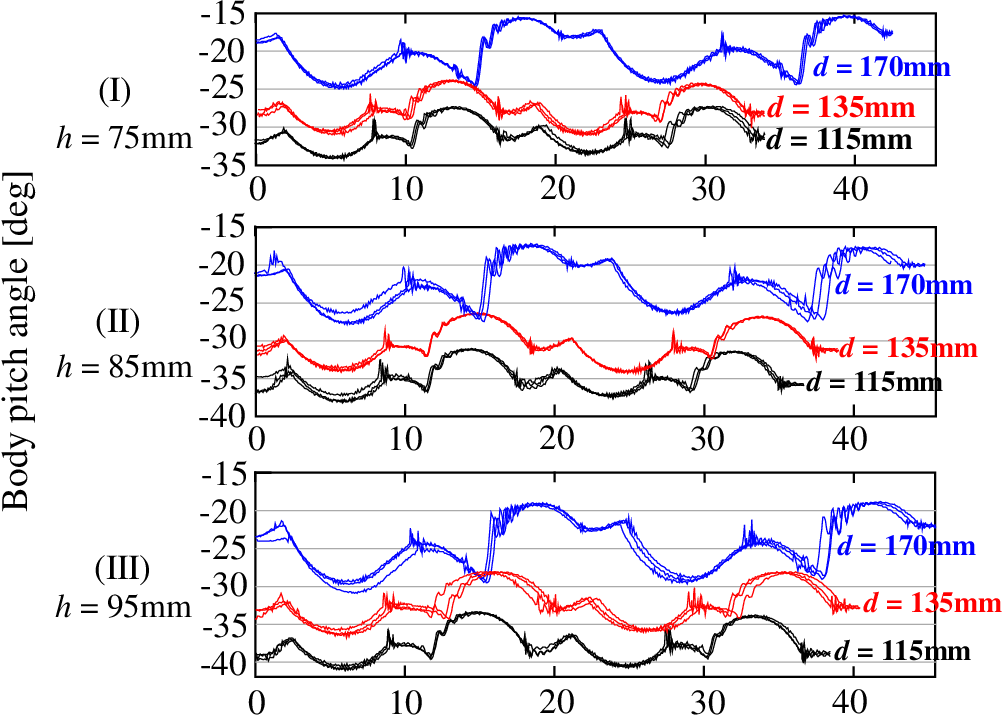}}
\vspace{-6mm}
\hspace{7mm}
  \caption{Results of stair climbing with various stair geometries. 
}
\label{fig:27_trials_of_climbing}
  \vspace{-5mm}
\end{figure}
Climbing experiments were performed on Stair I, II, and III defined in Table 
\ref{tab1}. The locomotions can be seen in https://youtu.be/2KtJypNziDs. 
Only at stair entry and exit, the seat position was changed, for reasons discussed later.
Except for the transient phases, the dummy and seat are fixed at the
analysis-derived optimal position, namely $X_5$ = -0.057 
m, -0.055 m, and -0.126 m for Stair I, II, and III, respectively.
Snapshots of steady stair ascent in Stair I and III are shown in
Figs. \ref{fig:snapshots}(a) and (b), respectively.
The 1st, 2nd, and 3rd images represent the moments at the beginning of
Mode 3, between Mode 3 and 4, and at the end of Mode 4, respectively.
Figures \ref{fig:graph}(a)-(c) depict data for one motion cycle on 
Stair I, II, and III, respectively, corresponding to one-step ascent achieved 
through Mode 3 to Mode 4.
In Figs. \ref{fig:graph}(a)-(c), row (I) shows the body pitch angle, with horizontal set to zero and forward tilt defined as positive.
Row (II) represents the wheelbase $l$ adjusted by the telescopic chassis. 
Row (III) shows the extension force estimated from measured current, where 
extension is defined as positive. 
Row (IV) plots the front- and rear-wheel torques computed from measured current, with forward rotation taken as positive.
Row (V) shows the required mechanical power of the chassis extension and 
retraction, calculated as $P_c = F(t)v(t)$, where $F(t)$ and $v(t)$ represent the 
extension force and velocity, respectively. 
Row (VI) shows the required mechanical power of each wheel pair, 
calculated as $P_w = \tau(t)\omega(t)$, where $\tau(t)$ and $\omega(t)$ denote 
the wheel torque and angular velocity, respectively. 
The remaining steps exhibit nearly identical time-series profiles.
\par
Figs. \ref{fig:graph}(a)-(c) show that, for all stairs, one motion cycle comprises 
locomotion Mode 3, a brief interval (Mode 0), and locomotion Mode 4, by which the wheelchair climbs one step.
As shown in row (II), for all stairs, the length of the telescopic chassis is 
controlled at constant velocities of approximately 0.02 m/s and -0.02 m/s during
Mode 3 and 4, respectively, according to the reference velocity defined by
Eq. (\ref{desired_velocity}), where $V_d$ = 0.02 m/s.
Throughout the cycle, the extension-retraction force is far lower than the 
rated force of 968 N, as shown in row (III).
Moreover, consistent with contribution (C), the maximum extension-retraction 
force shows little variation despite the increase in stair steepness from Stair 
I to II to III.
From row (IV), it can be seen that, despite the stair steepness, the wheels 
are driven with small torque (the rated torque is 3.42 Nm), sufficient not to interfere with the body extension motion, demonstrating contribution (B).
Note that, as described in Sec. \ref{Operation and drive method}, the 
commanded wheel torque magnitude was fixed at 0.3 Nm. However, as shown 
in row (IV), the actual output torque varied mainly due to the back 
electromotive force. Nevertheless, the output torque magnitude remained generally close to the commanded value.
\par
When comparing (V) with (VI) in Fig. \ref{fig:graph}, the required mechanical power of the chassis is 
significantly larger than that of the wheels (Note that the values on the vertical axes are different). 
Based on the results, 
mechanical work performed per unit vertical height
was calculated for Stair I, II, and III, resulting in 284 J/m
(267 J/m for the telescopic chassis and 17 J/m for both wheels),
289 J/m (274 J/m for the chassis and 15 J/m for both wheels),
and 203 J/m (192 J/m for the chassis and 11 J/m for both wheels), respectively.
Therefore, for all stairs, the contribution of the telescopic chassis to the total 
mechanical work was greater than 94\%, which experimentally confirms contribution (B).
Furthermore, counterintuitively, Stair III requires less mechanical work than Stair I, experimentally verifying contribution (C).
\par
%
Fig. \ref{fig:dummy-3kg-and-11kg} shows the body pitch angle over two cycles during ascent of Stair 
III with dummy masses of 3 kg and 11 kg. By setting the seat position 
appropriately for each dummy mass, stable stair ascent was achieved regardless of the payload weight.
In addition, stair-climbing experiments were conducted on nine different 
staircases constructed from all combinations of the three riser heights 
and three tread depths used for Stairs I-III listed in Table \ref{tab1}, 
including staircases with the Blondel rule values of $2h + d$ ranging 
from 53 cm to 72 cm, covering even rare staircase geometries encountered in practice. Each 
staircase was traversed three times, resulting in a total of 27 
consecutive trials. As shown in Fig. \ref{fig:27_trials_of_climbing}, 
the body pitch angle over two cycles was nearly identical for each 
staircase, demonstrating robust stair-climbing capability without any 
failures in the 27 consecutive trials.
\par
When the rear wheels attempt to ascend the first step from level
ground, the body inclination is zero. Therefore, when the center of mass is 
located toward the rear of the vehicle, the rear wheels are unable to climb the first step.
Accordingly, the seat was initially placed close to the front
wheels until the wheelchair transitioned into a steady stair-climbing inclination.
\par
\section{Conclusion and future work}
\label{Discussion and Conclusion}
In this paper, we propose a simple locomotion mechanism that enables gradual and 
robust stair ascent by relying on chassis extension and retraction,
without adding specialized mobility mechanisms such as crawlers.
Currently, seat translation is controlled remotely by an operator. Since 
a strong correlation exists between the stair inclination (body pitch 
angle) and the seat position required for stable stair climbing, future 
versions of the system will automatically adjust the seat position 
according to the measured body inclination. 　
In addition, the seat position must be controlled within a safe range that 
prevents both tip-over and slippage of the stationary wheels remaining on 
the tread. Therefore, load cells will be integrated into the wheel axles to estimate the normal force acting on each wheel. 
The allowable seat-travel range will be determined so that all four wheels maintain sufficiently large normal forces for static stability, while the stationary wheels simultaneously sustain enough normal force to provide the friction necessary for reliable traction on the stair treads.
\par
The proposed automatic seat-position control ensures stability while allowing the user to focus on telescopic-body operation. However, further improvements in operability are needed.
As shown in Fig. \ref{fig:diagram_climbing_stairs}, although the chassis must be extended or retracted until the rolling 
wheels are sufficiently positioned on the next tread, it is difficult 
for an occupant with limited visibility of the wheels to stop the motion before the wheels hit the next riser.
To address this issue, an automatic motion-limiting function will be introduced.
Specifically, the horizontal reaction force acting on each wheel from 
the riser will be estimated using the aforementioned load cells and the body inclination angle. Once contact with the next riser is detected, the telescopic motion will be automatically stopped. As a result, the occupant only needs to extend or retract the body until the imposed limit is reached, thereby simplifying the stair-climbing operation.
\par
In addition, as described in Sec. \ref{Analysis and discussion}, by reciprocating the seat position between locomotion Mode 3 and 4
(Table \ref{tab2}), improved stair-climbing performance is expected even on
slippery surfaces caused by rain or freezing.
The highest slip risk in Mode 3 and 4 occurs at the instant when the
wheel starts climbing the stair edge as shown in Figs. 
\ref{fig:free_body_diagram}(a) and (b).
Therefore, initiating seat motion toward the next target position immediately
after the moments of Figs. \ref{fig:free_body_diagram}(a) and (b) and 
continuing seat sliding with body extension and retraction will not result in significant time loss.
The aforementioned load-cell-based seat safety control is expected to be effective for this purpose.
\par

\end{document}